\documentclass{article}
\usepackage{spconf,amsmath,graphicx,hyperref}

\usepackage{graphicx}
\usepackage{xcolor}
\usepackage{booktabs}
\usepackage{multirow}
\usepackage{amsmath}
\usepackage{lipsum}
\usepackage{xcolor}
\usepackage{verbatim}
\usepackage[dvipsnames]{xcolor}
\usepackage{array} 
\usepackage{makecell}
\usepackage[table]{xcolor}
\usepackage{xcolor}
\definecolor{palegreen}{rgb}{0.91, 0.96, 0.88}
\usepackage[most]{tcolorbox}
\usepackage{booktabs}
\usepackage{graphicx}
\usepackage[table,xcdraw]{xcolor}
\usepackage{listings} 
\usepackage{makecell}
\usepackage{algorithm}
\usepackage{algorithmic}
\usepackage{amsmath}
\usepackage{amsfonts} 
\usepackage{amssymb}
\usepackage{dsfont}
\usepackage{pifont}
\usepackage{enumitem}

\usepackage{xcolor}
\definecolor{darkerblue}{RGB}{110, 145, 195} 
\definecolor{darkerred}{RGB}{180, 125, 120} 
\definecolor{darkeryellow}{RGB}{205, 180, 90} 
\definecolor{darkerorange}{RGB}{215, 160, 100} 
\definecolor{darkerpurple}{RGB}{130, 100, 160}

\newcommand{\green}[1]{\textcolor{ForestGreen}{\textbf{#1}}}
\newcommand{\red}[1]{\textcolor{red}{\textbf{#1}}}

\newcommand{\num}[2]{$\displaystyle {#1}_{\textcolor{gray}{\textit{#2}}}$} 
\definecolor{green1}{HTML}{A8E6CF}
\definecolor{green2}{HTML}{D4F5C7}
\definecolor{red1}{HTML}{E8A1A1}
\definecolor{red2}{HTML}{F5BDBD}
\definecolor{yellow1}{HTML}{FFF7C2} 
\definecolor{orange1}{HTML}{FFE4B2}
\definecolor{blue1}{HTML}{C9E4F6}
\definecolor{purple1}{HTML}{D9C9EB}
\definecolor{pink1}{HTML}{F4C6D7}

\title{Z-Scores: A Metric for Linguistically Assessing Disfluency Removal}
\name{Maria Teleki, Sai Janjur, Haoran Liu, Oliver Grabner, Ketan Verma, Thomas Docog, Xiangjue Dong, Lingfeng Shi, Cong Wang, Stephanie Birkelbach, Jason Kim, Yin Zhang, James Caverlee}
\address{Texas A\&M University}

\begin{document}

\maketitle

\begin{abstract}
Evaluating disfluency removal in speech requires more than aggregate token-level scores. Traditional word-based metrics such as precision, recall, and F1 ($\mathcal{E}$-Scores) capture overall performance but cannot reveal why models succeed or fail. We introduce $\mathcal{Z}$-Scores, a span-level linguistically-grounded evaluation metric that categorizes system behavior across distinct disfluency types (EDITED, INTJ, PRN). Our deterministic alignment module enables robust mapping between generated text and disfluent transcripts, allowing $\mathcal{Z}$-Scores to expose systematic weaknesses that word-level metrics obscure. By providing category-specific diagnostics, $\mathcal{Z}$-Scores enable researchers to identify model failure modes and design targeted interventions -- such as tailored prompts or data augmentation -- yielding measurable performance improvements. A case study with LLMs shows that Z-scores uncover challenges with INTJ and PRN disfluencies hidden in aggregate F1, directly informing model refinement strategies. 
\end{abstract}


\section{Introduction}
\label{sec:intro}

\begin{figure}
    \centering
    \includegraphics[width=1\linewidth]{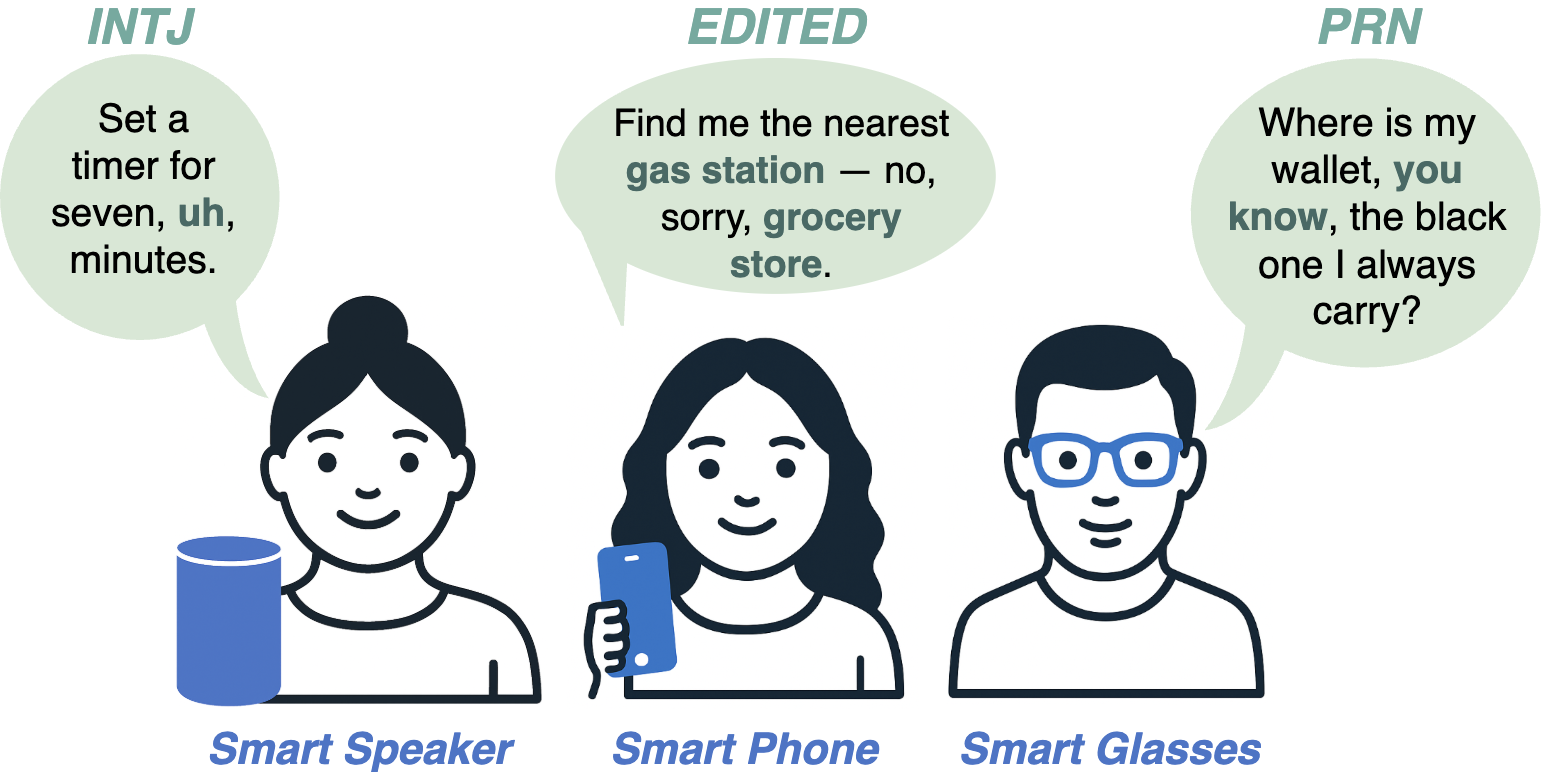}
    \caption{Removing disfluencies such as INTJ (\textit{uh}), EDITED (\textit{gas station} is replaced with \textit{grocery store}), and PRN (\textit{you know}) ensures clean text input for downstream tasks. Our proposed $\mathcal{Z}$-Score metric uncovers categorical errors, which can drive targeted model improvements.}
    \label{fig:Fig1}
\end{figure}

Spontaneous speech is filled with disfluencies systematically categorized in Shriberg's linguistic framework~\cite{Shriberg.1994}, which defines \textbf{interjections (INTJ)} -- e.g., \textit{um}, \textit{uh}, \textit{uh-huh} -- signaling hesitation or backchanneling; \textbf{parentheticals (PRN)} -- e.g., \textit{you know}, \textit{I mean} -- functioning as meta-commentary or discourse markers; and \textbf{edited nodes (EDITED)} -- e.g., \textit{Where did I put my keys -- sorry, phone?} -- capturing false starts, repairs, and restarts. 
As illustrated in Figure \ref{fig:Fig1}, interactions with smart speakers or wearable devices often contain such disfluencies, which have been shown to degrade downstream task performance in transcription, translation, and conversational recommendation \cite{teleki25_horror, retkowski2025summarizingspeechcomprehensivesurvey}.

Existing approaches for evaluating disfluency removal systems rely primarily on word-level precision, recall, and F1 scores. While these metrics are useful for coarse performance measurement, they provide only a limited view. Crucially, they cannot explain why a system succeeds or fails. For example, a model’s overall F1 may look strong, yet it may consistently fail to remove parentheticals or interjections — weaknesses that remain hidden in aggregate scores.

To address this gap, we introduce $\mathcal{Z}$-Scores, a span-level diagnostic metric for disfluency removal that can be used to drive modeling improvements. $\mathcal{Z}$-Scores quantify how models handle distinct categories of disfluencies (EDITED, INTJ, PRN) by leveraging a deterministic alignment module that ensures reliable mapping between generated outputs and disfluent transcripts. This enables evaluation at the level of linguistic phenomena, not just tokens, and makes it possible to uncover systematic behaviors that traditional metrics obscure.

$\mathcal{Z}$-Scores complement existing word-level metrics with linguistic interpretability, as they let us evaluate how well models remove these distinct disfluency types rather than collapsing them into a single F1 score. $\mathcal{Z}$-Scores empower researchers and practitioners to ask category-aware questions such as: \textit{Do generative models remove interjections as effectively as parentheticals? Do models fail to remove parentheticals when they occur near certain linguistic phenomena? How do prompting strategies shift performance across disfluency types?} \textbf{By providing a diagnostic lens, $\mathcal{Z}$-Scores serve as a bridge between computational evaluation and linguistic analysis, empowering researchers to uncover hidden weaknesses and guide modeling improvements in disfluency removal. We release an open-source Python package at \url{https://github.com/mariateleki/zscore}.}

\begin{figure*}[t]
    \centering
    \includegraphics[width=0.81\linewidth]{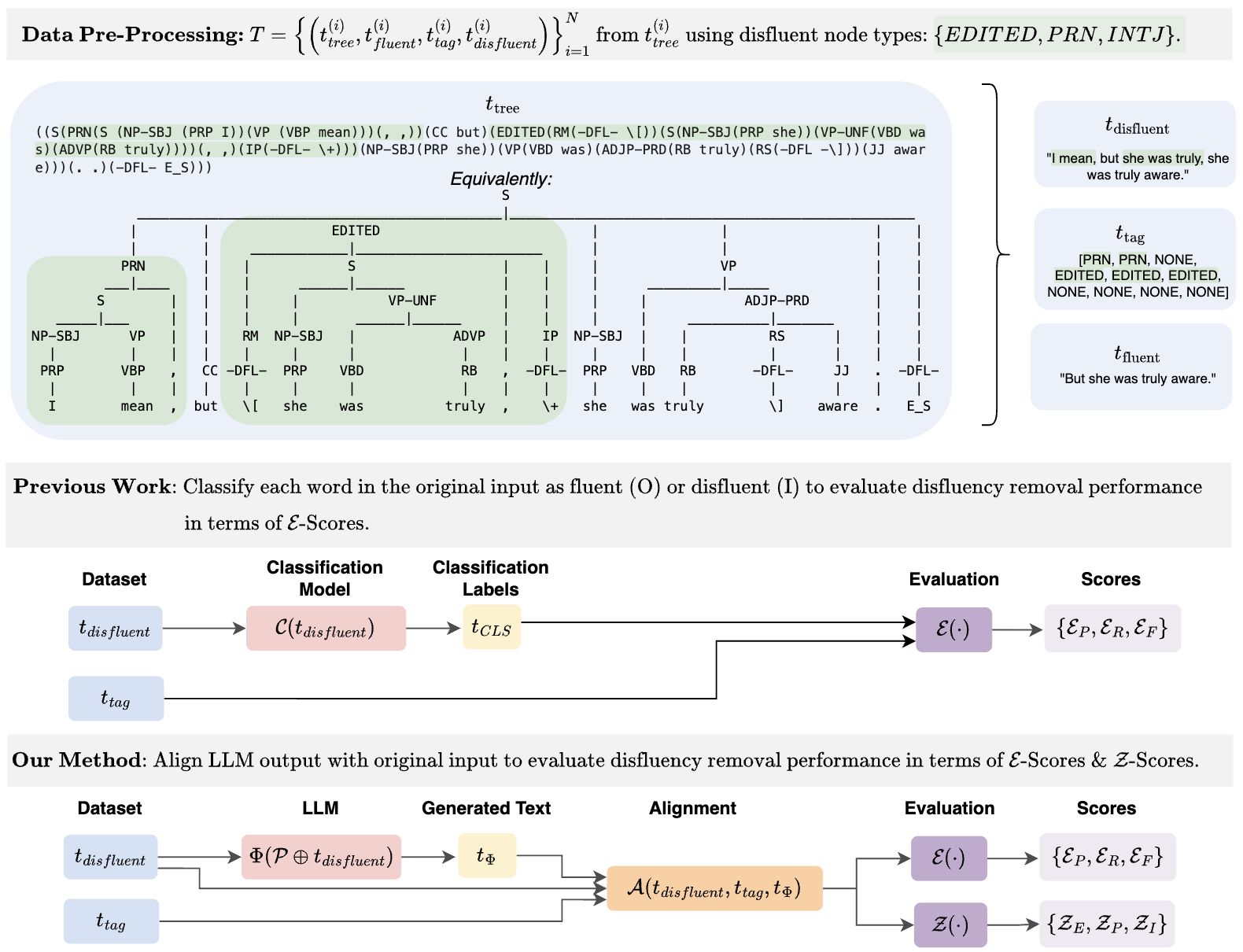}
    \caption{\textbf{$\mathcal{Z}$-Score Framework}: Previous work treated the disfluency removal task as a \textcolor{darkerred}{classification} task. In contrast, we design an \textcolor{darkerorange}{alignment module, $\mathcal{A}$}, to allow GMs to be used for the disfluency removal task. This module aligns generated text with disfluent text to enable the \textcolor{darkerpurple}{$\mathcal{E}$- and $\mathcal{Z}$-scoring} of \textcolor{darkerred}{GMs, $\Phi$}, allowing category-specific performance analysis and supporting targeted modeling strategies.}
    \label{fig:ExperimentalFramework}
\end{figure*}

\section{Relation to Prior Work}
\label{sec:prior}

Previous work evaluates performance in terms of word-based precision, recall, and F-score -- we denote these as $\mathcal{E}$-Scores -- for the disfluency removal task as shown in Figure \ref{fig:ExperimentalFramework}. To calculate $\mathcal{E}$-Scores, an alignment between the original text and the model-generated text is required, as shown in Table \ref{tab:alignment}. Hence, previous work in this task has primarily been formulated as sequence classification, because obtaining the alignment between the ground-truth disfluent text and the GM-generated text is difficult. In our work, we provide a method for obtaining this alignment (\S \ref{sec:alignment}), enabling the use of generative language models for the disfluency removal task.

\subsection{Generative Models (GMs)}

Compared to prior specialized models, generative models (GMs) like LLMs and SLMs offer two distinct advantages for this task: First, their extensive pretraining on diverse internet-scale text corpora provides rich world knowledge and a deep semantic understanding that smaller, task-specific models lack. This allows them to better interpret domain-specific vocabulary and contextual nuances within disfluent speech. Second, their development cycle allows for continuous updates to the base models, enabling them to adapt to evolving language -- including new slang and concepts -- thereby addressing the challenge of temporal language shift. As shown in Figure \ref{fig:ExperimentalFramework}, previous work has focused on using generative models as part of the classification model pipeline for the disfluency removal task, either (i) using GM as encoders, or (ii) using GM to generate synthetic disfluency data for fine-tuning; we detail the use of GMs in these ways next. The use of generative models for disfluency removal has been underexplored due to a lack of methodology to align generated text with disfluent text. To enable this alignment, we develop a new alignment module, $\mathcal{A}$ (\S \ref{sec:alignment}). 

\subsection{Classification Models}
Classification models perform token-by-token classification of disfluent text, labeling each token in the sequence as either disfluent, $I$ (``inside'' the disfluent region) or fluent, $O$ (``outside'') as shown in column $t_{CLS}$ of Table \ref{tab:alignment}.
We detail the major approaches:
(i) Some works propose joint encoding.
\cite{wagner2024large} use a joint LLM-ASR architecture.
\cite{Wang_Wang_Che_Zhao_Liu_2021} and \cite{Dong_Wang_Yang_Chen_Xu_Xu_2019} propose transformer variants with joint encoding of labeled and unlabeled input sequences. 
(ii) Many works perform synthetic disfluency insertion on the original text before fine-tuning. 
\cite{ding2024disfluency} and \cite{cheng2024boosting} generate synthetic disfluency data from OpenAI LLMs and fine-tune small BERT models.
\cite{Wang_Wang_Che_Liu_2020} and \cite{chaudhury-etal-2024-dacl-disfluency} rely on rule-based methods for generating synthetic disfluency data.
(iii) \cite{jamshid-lou-johnson-2020-improving} proposes a span classifier for building parse trees. 
(iv) Earlier works use methods such as LSTM, transition modeling, and noisy channel modeling for sequence classification \cite{Yang_Yang_Ma_2020, Lou_Wang_Johnson_2019, bach2019noisy, wang2018semi, Lou_Johnson_2017, zayats2016disfluency, johnson-charniak-2004-tag, Charniak.2001, Hindle_1983}.
Additionally, some of these models force alignment with the original sequence via constrained decoding. While this method enables calculation of $\mathcal{E}$-Scores, it can also reduce the expressive power of the GM by reducing the search space during decoding.

\subsection{The Alignment Problem} 
Two previous works have attempted to address the problem of aligning generated text with the original disfluent text.
(i) \cite{chaudhury-etal-2024-dacl-disfluency} perform this alignment with \textit{longest common subsequence (LCS)}. However, this approach is suboptimal, as we will show in \S \ref{sec:alignment}.
In contrast, \cite{Salesky_Sperber_Waibel_2019} -- who also use a translation-based model -- avoid the alignment, instead evaluating their performance using the BLEU and ROUGE metrics. These types of n-gram-based metrics ignore alignment, meaning the fine-grained parse tree information about where specific disfluencies are removed is lost. 
(ii) \cite{jamshid-lou-johnson-2020-end} (who also point out this problem with the evaluation of \cite{Salesky_Sperber_Waibel_2019}) propose a statistical weighting method based on Gestalt pattern matching \cite{gestalt} to align the generated text with the original disfluent text. However, this method is not deterministic, and therefore lacks the alignment guarantee that we are able to obtain with our alignment method (detailed in \S \ref{sec:alignment}).

\begin{table}
    \centering
    \resizebox{0.9\columnwidth}{!}{%
    \begin{tabular}{@{}llll|llllll@{}}
        \toprule
        \textcolor{darkerblue}{$t_{disfluent}$} & \textcolor{darkerblue}{$t_{tag}$} & \textcolor{darkeryellow}{$t_\Phi$} & \textcolor{darkeryellow}{$t_{CLS}$}
        & $\mathds{1}_{gt}$ & $\mathds{1}_{pred}$ 
        & $\mathds{1}_{tp}$ & $\mathds{1}_{tn}$ & $\mathds{1}_{fp}$ & $\mathds{1}_{fn}$ 
        \\
        \midrule
            i &   PRN  &    i  &    $I$   &  $1$     &    $0$   &    $0$   &    $0$   &    $0$   &    $1$ \\
         mean &   PRN  & mean  &    $I$   &  $1$     &    $0$   &    $0$   &    $0$   &    $0$   &    $1$ \\
          but &  NONE  &  but  &    $O$   &  $0$     &    $0$   &    $0$   &    $1$   &    $0$   &    $0$ \\
          she & EDITED &       &    $I$   &  $1$     &    $1$   &    $1$   &    $0$   &    $0$   &    $0$ \\
          was & EDITED &       &    $I$   &  $1$     &    $1$   &    $1$   &    $0$   &    $0$   &    $0$ \\
        truly & EDITED &       &    $I$   &  $1$     &    $1$   &    $1$   &    $0$   &    $0$   &    $0$ \\
          she &   NONE &       &    $I$   &  $0$     &    $1$   &    $0$   &    $0$   &    $1$   &    $0$ \\
           -  &     -  & Luna &    $O$   &  $*$     &    $*$   &    $*$   &    $*$   &    $*$   &    $*$ \\
          was &   NONE &   was &    $O$   &  $0$     &    $0$   &    $0$   &    $1$   &    $0$   &    $0$ \\
        truly &   NONE & truly &    $O$   &  $0$     &    $0$   &    $0$   &    $1$   &    $0$   &    $0$ \\
        aware &   NONE & aware &    $O$   &  $0$     &    $0$   &    $0$   &    $1$   &    $0$   &    $0$ \\
        \bottomrule
    \end{tabular}
    }
    \caption{\textbf{Alignment Example}: Our method, \textcolor{darkerorange}{$\mathcal{A}$}, is able to align \textcolor{darkerblue}{$t_{disfluent}$} and \textcolor{darkeryellow}{$t_\Phi$} for \textcolor{darkerpurple}{$\mathcal{E}$- and $\mathcal{Z}$-scoring}, in contrast to previous methods which simply perform sequence classification (\textcolor{darkeryellow}{$t_{CLS}$}) to allow scoring. Hallucinated tokens (*) are filtered before scoring, ensuring they do not affect results, allowing continuity with previous evaluation methodology.}
    \label{tab:alignment}
\end{table}

\section{$\mathcal{Z}$-Score Framework}

Figure \ref{fig:ExperimentalFramework} shows our framework for evaluating GMs on the disfluency removal task.  We introduce $\mathcal{Z}$-scoring, a span-level metric for quantifying the types (i.e. EDITED, INTJ, PRN) of disfluencies each model is able to remove. We design an alignment module to enable both traditional $\mathcal{E}$-scoring and this span-level $\mathcal{Z}$-scoring for GMs.

\subsection{Alignment Module ($\mathcal{A}$)}\label{sec:alignment}

Prior alignment strategies such as LCS, BLEU/ROUGE-based evaluation, and statistical weighting methods either produce systematic errors (e.g., LCS preserves disfluent tokens), fail to capture span-level disfluencies (BLEU/ROUGE), or lack determinism (statistical weighting). In contrast, our deterministic alignment module $\mathcal{A}$ ensures reliable alignment.

The alignment module is responsible for constructing the $t_{disfluent}$, $t_{tag}$, $t_\Phi$ columns which are used by $\mathcal{E}(\cdot)$ and $\mathcal{Z}(\cdot)$.
Hence, we first tokenize using TreebankWordTokenizer. We then use a variation of \textbf{Gestalt matching \cite{gestalt}, $\mathcal{G}$}, to align the tokens in the generated, $t_\Phi$, and ground-truth, $t_{disfluent}$, sequences for comparison. $t_{tag}$ is metadata for $t_{disfluent}$, as it contains sequence information pertaining to the set of disfluent tags: \{EDITED, PRN, INTJ\}.

However, $\mathcal{G}$ cannot be straightforwardly applied, as it performs \textit{early matching} -- e.g. for input tuples in the form $(t_{disfluent}, t_{tag}, t_{\Phi})$, 
\red{$\mathcal{G}$ incorrectly yields}:

\begin{equation*}
\begin{split}
[(\text{the},\text{\textbf{EDITED}},\text{\textbf{the}}),
(\text{the},\text{NONE},\emptyset),
(\text{cat},\text{NONE},\emptyset)]
\end{split}
\end{equation*}

To fix this problem, we modify $\mathcal{G}$ to create $\mathcal{A}$ for alignment. We append a special token and the tag (e.g. \textit{``the§EDITED''}) to disfluent tokens to form $t_\Phi'$, before running $\mathcal{G}(t_{disfluent},t_\Phi')$. This step forces disfluent tokens into the \textit{replace} case of $\mathcal{G}$, where we match to valid NONE tags before valid disfluent tags. 
Hence,
\green{\textsc{$\mathcal{A}$} correctly yields}:
\begin{equation*}
\begin{split}
[(\text{the},\text{EDITED},\emptyset),
(\text{the},\text{\textbf{NONE}},\text{\textbf{the}}),
(\text{cat},\text{NONE},\emptyset)]
\end{split}
\end{equation*}

We show an example of this alignment in Table \ref{tab:alignment}, using the example from Figure \ref{fig:ExperimentalFramework}. The character $*$ marks hallucinated tokens produced by the GM, which we treat as an artifact of GMs and explicitly remove during post-processing,\footnote{Hence, $\mathcal{A}$ also facilitates a straightforward filtering step that eliminates hallucinated tokens entirely from the final system output.} ensuring that this content does not propagate to downstream tasks. 

\subsection{$\mathcal{E}$-Scores} \label{sec:e-scores}

The $\mathcal{E}$-Score function:
\[
\mathcal{E}(\mathcal{A}(t_{disfluent}, t_{tag}, t_{\Phi})) \rightarrow 
\{ \mathcal{E}_{F}, \mathcal{E}_{P}, \mathcal{E}_{R}\}
\]
returns a set of scores for the word-based F1, precision, and recall scores for the disfluency removal task using the $\mathcal{A}$ alignment. Previous work measured performance in terms of $\mathcal{E}$-Scores. Looking to Table \ref{tab:alignment}, $\mathcal{E}$-Scores can only be calculated given a correct alignment between $t_{disfluent}$ and $t_\Phi$, which we obtain with our alignment module, $\mathcal{A}$. We demonstrate calculating $\mathcal{E}$-Scores with Table \ref{tab:alignment}. From the alignment $\mathcal{A}$, $\mathds{1}_{gt}$ and $\mathds{1}_{pred}$ are calculated by comparing $t_{\Phi}$ to $t_{tag}$: 

\begin{itemize}[left=0pt, noitemsep, topsep=0pt]
    \item $\mathds{1}_{gt}$:  \textit{Based on the ground truth parse tree, should this word be removed by $\Phi$?}
    \item $\mathds{1}_{pred}$: \textit{Was this word actually removed by $\Phi$?}
\end{itemize}

Then, from $\mathds{1}_{\{tp,fn,tn,fp\}}$, 
the $\mathcal{E}$-Scores are calculated:

\begin{itemize}[left=0pt, noitemsep, topsep=0pt]
    \item $\mathcal{E}_P=\frac{\Sigma_{tp}}{\Sigma_{tp}+\Sigma_{fp}}=\frac{3}{3+1} \to 75.0$
    \item $\mathcal{E}_R=\frac{\Sigma_{tp}}{\Sigma_{tp}+\Sigma_{fn}}=\frac{3}{3+2} \to 60.0$
    \item $\mathcal{E}_F=\frac{2\cdot\mathcal{E}_P\cdot\mathcal{E}_R}{\mathcal{E}_P+\mathcal{E}_R}=\frac{2\cdot0.75\cdot0.60}{0.75+0.60} \to 66.0$
\end{itemize}

\underline{$\mathcal{E}_P$ penalizes over-deletion}, hence $\mathcal{E}_P$  is low when fluent tokens are incorrectly removed.
Recall that $fp$ indicates $\Phi$ should not have removed the token, but did.
In contrast, 
\underline{$\mathcal{E}_R$ penalizes under-deletion}, hence $\mathcal{E}_R$ is low when ground-truth disfluencies remain in the output.
Recall that $fn$ indicates $\Phi$ should have removed the token, but didn't. 
Then, \underline{$\mathcal{E}_F$ is the harmonic mean of $\mathcal{E}_P$ and $\mathcal{E}_R$} and the main indicator of overall disfluency removal performance; it is low if either $\mathcal{E}_P$ or $\mathcal{E}_R$ is low.

\subsection{$\mathcal{Z}$-Scores} \label{sec:z-scores}

We propose $\mathcal{Z}$-Scores, a new span-level metric for the disfluency removal task. $\mathcal{Z}$-Scores are more informative in terms of model performance on individual disfluency types. We use this new metric to examine the type of disfluencies each model removes well, or struggles to remove. The $\mathcal{Z}$-Score function:
\[
\mathcal{Z}(\mathcal{A}(t_{disfluent}, t_{tag}, t_{\Phi})) \rightarrow 
\{ \mathcal{Z}_{E}, \mathcal{Z}_{I}, \mathcal{Z}_{P}\}
\]
returns a set of scores for the percentage of EDITED, PRN, and INTJ  nodes that the model was successfully able to remove using the $\mathcal{A}$ alignment:

\begin{itemize}[left=0pt, noitemsep, topsep=0pt]
    \item 
    $\mathcal{Z}_E
    = 
    \frac
    {\mathds{1}_{gt} \land (w_{tag} =  EDITED) \land \mathds{1}_{pred}}
    {\mathds{1}_{gt}  \land (w_{tag} =  EDITED)}
    =
    \frac
    {3}
    {3}
    \to
    100\%
    $
    \item 
    $\mathcal{Z}_I
    = 
    \frac
    {\mathds{1}_{gt} \land (w_{tag} =  INTJ) \land \mathds{1}_{pred}}
    {\mathds{1}_{gt}  \land (w_{tag} =  INTJ)}
    =
    \frac
    {0}
    {0}
    \to
    \textit{NaN}
    $
    \item
    $\mathcal{Z}_P
    = 
    \frac
    {\mathds{1}_{gt} \land (w_{tag} =  PRN) \land \mathds{1}_{pred}}
    {\mathds{1}_{gt}  \land (w_{tag} =  PRN)}
    =
    \frac
    {0}
    {2}
    \to
    0\%
    $
\end{itemize}

Because $t_{tag}$ is constructed using a top-down recursive approach, the tags are span-level. Hence, $\mathcal{Z}$-Scores can be considered to be a span-level metric, whilst $\mathcal{E}$-Scores can be considered to only be a word-level metric.

\begin{table}[]
\centering
\resizebox{0.99\columnwidth}{!}{%
\begin{tabular}{@{\hspace{6pt}}>{\columncolor{gray!20}}l>{\columncolor{gray!20}}llll||lll@{}}
 \toprule
 \rowcolor{gray!5}
 \multicolumn{1}{l}{}
 & \multicolumn{1}{l}{}
 & \multicolumn{6}{c}{\texttt{gpt-4o-mini}} 
\\
\midrule
 \rowcolor{gray!5}
  \multicolumn{1}{c}{$M$} & 
  & 
  \multicolumn{1}{c}{$\mathcal{E}_F$} & \multicolumn{1}{c}{$\mathcal{E}_P$} & \multicolumn{1}{c}{$\mathcal{E}_R$} & \multicolumn{1}{c}{$\mathcal{Z}_E$} & \multicolumn{1}{c}{$\mathcal{Z}_I$} & \multicolumn{1}{c}{$\mathcal{Z}_P$}  \\
  \midrule
$s$ & $P_0$
& \cellcolor{red2}\num{72.69}{5.79} & \num{75.61}{7.05} & \num{70.48}{7.35} & \num{85.20}{8.23} & \num{61.89}{11.08} & \num{65.02}{20.99}\\
$s$ & $P_1$
& \cellcolor{green2} \num{81.94}{3.75} & \num{84.47}{4.92}& \num{79.90}{5.65}& \cellcolor{yellow1}\num{83.67}{9.27}& \num{78.28}{8.10}& \cellcolor{orange1}\num{74.86}{22.06} \\
$s$ & $P_2$
& \cellcolor{green2}\num{79.86}{5.42}& \num{76.88}{7.02} & \num{83.52}{6.12}& \cellcolor{yellow1}\num{87.45}{7.48}& \cellcolor{orange1}\num{79.60}{8.89}& \num{87.09}{15.46} \\
\bottomrule
\end{tabular}%
}
\caption{\textit{\textbf{Metaprompting} (\num{\textit{mean}}{\textit{std. dev}}): Incorporating short prompts with common disfluencies ($\mathcal{P}_1,\mathcal{P}_2$) improves performance. While $\mathcal{E}$-Scores suggest modest overall gains, $\mathcal{Z}$-Scores reveal that these improvements are primarily driven by better INTJ and PRN removal, highlighting the \textbf{diagnostic value} of our proposed metric.}}
\label{tab:metaprompting}
\end{table}

\section{A Case Study: Metaprompting}

We conduct a small-scale study with metaprompting~\cite{zhang2023meta} to illustrate the utility of our proposed metric. Experiments are conducted on the Switchboard dataset \cite{godfrey1992switchboard}, though our method is generalizable to other corpora. We use gpt-4o-mini as a representative GM, with results shown in Table~\ref{tab:metaprompting}. 

We start with our baseline prompt for disfluency removal, $\mathcal{P}_0$. $\mathcal{P}_0$ achieves reasonable overall $\mathcal{E}$-Scores, but $\mathcal{Z}$-Scores reveal clear weaknesses: the model handles EDITED disfluencies well ($\mathcal{Z}_E$=85.20), yet performs poorly on INTJ (interjections such as \textit{uh}, \textit{um}, with $\mathcal{Z}_I$=61.89) and PRN (parentheticals such as \textit{I mean}, with $\mathcal{Z}_P$=65.02). These modeling deficiencies are hidden when looking only at $\mathcal{E}$-Scores.

In comparison, metaprompts $\mathcal{P}_1$ and $\mathcal{P}_2$ include explicit examples of INTJ and PRN disfluencies. For these prompts, the $\mathcal{Z}$-Scores show marked improvements: $\mathcal{Z}_I$ increases approximately 16 points, while $\mathcal{Z}_P$ rises approximately 9 points. In contrast, $\mathcal{Z}_E$ remains stable, confirming that the gains stem specifically from better handling of INTJ and PRN.

Importantly, $\mathcal{Z}$-Scores make this diagnostic insight possible. \textbf{Whereas $\mathcal{E}$-Scores provide only an aggregate view of precision and recall, $\mathcal{Z}$-Scores reveal that performance gains are localized to the specific linguistic phenomenon of INTJ and PRN disfluencies.} This illustrates the value of $\mathcal{Z}$-Scores as \underline{a fine-grained diagnostic tool for understanding} \underline{model behavior and driving model improvements.} 

Future directions include the development of category-specific prompting schemes, architectural innovations (e.g. disfluency category specialized adapters), and tailored augmentation pipelines, with $\mathcal{Z}$-Scores serving as the central mechanism for diagnosis-driven model refinement.

\section{Conclusion}
\label{sec:conclusion}

We introduced $\mathcal{Z}$-Scores, a span-level evaluation metric that complements traditional $\mathcal{E}$-Scores for disfluency removal with linguistic assessment based on specific disfluency categories (EDITED, INTJ, PRN). A case study with LLMs illustrates how $\mathcal{Z}$-Scores reveal differences in handling specific disfluency types (INTJ, PRN) and how $\mathcal{Z}$-Scores can be used to drive model performance improvements via targeted prompting. We release our metric as an open source Python package, providing the community with a standardized resource for future research.

\bibliographystyle{IEEEbib}
\bibliography{strings,refs}

{\small © 2025 IEEE. Personal use of this material is permitted. Permission from IEEE must be obtained for all other uses, in any current or future media, including reprinting/republishing this material for advertising or promotional purposes, creating new collective works, for resale or redistribution to servers or lists, or reuse of any copyrighted component of this work in other works.}

\end{document}